\documentclass[10pt,twocolumn,letterpaper]{article}

\usepackage{cvpr}
\usepackage{times}
\usepackage{epsfig}
\usepackage{graphicx}
\usepackage{amsmath}
\usepackage{amssymb}

\usepackage[breaklinks=true,bookmarks=false]{hyperref}

\cvprfinalcopy 

\begin{document}

\title{Im2Avatar: Colorful 3D Reconstruction from a Single Image} 

\author{Yongbin Sun\\
MIT\\
{\tt\small yb\_sun@mit.edu}
\and
Ziwei Liu\\
UC Berkley\\
{\tt\small zwliu@icsi.berkeley.edu}
\and
Yue Wang\\
MIT\\
{\tt\small yuewang@csail.mit.edu}
\and
Sanjay E. Sarma\\
MIT\\
{\tt\small sesarma@mit.edu}
}
\maketitle

\begin{abstract}


Existing works on single-image 3D reconstruction mainly focus on shape recovery.
In this work, we study a new problem, that is, simultaneously recovering 3D shape and surface color from a single image, namely ``colorful 3D reconstruction''.
This problem is both challenging and intriguing because the ability to infer textured 3D model from a single image is at the core of visual understanding.
Here, we propose an end-to-end trainable framework, Colorful Voxel Network (CVN), to tackle this problem.
Conditioned on a single 2D input, CVN learns to decompose shape and surface color information of a 3D object into a 3D shape branch and a surface color branch, respectively.
Specifically, for the shape recovery, we generate a \textit{shape volume} with the state of its voxels indicating occupancy.
For the surface color recovery, we combine the strength of appearance hallucination and geometric projection by concurrently learning a \textit{regressed color volume} and a \textit{2D-to-3D flow volume}, which are then fused into a \textit{blended color volume}. 
The final textured 3D model is obtained by sampling color from the blended color volume at the positions of occupied voxels in the shape volume.
To handle the severe sparse volume representations, a novel loss function, Mean Squared False Cross-Entropy Loss (MSFCEL), is designed.
Extensive experiments demonstrate that our approach achieves significant improvement over baselines, and shows great generalization across diverse object categories and arbitrary viewpoints.

\end{abstract}

\section{Introduction}

\begin{figure}[t]
\includegraphics[width=0.45\textwidth]{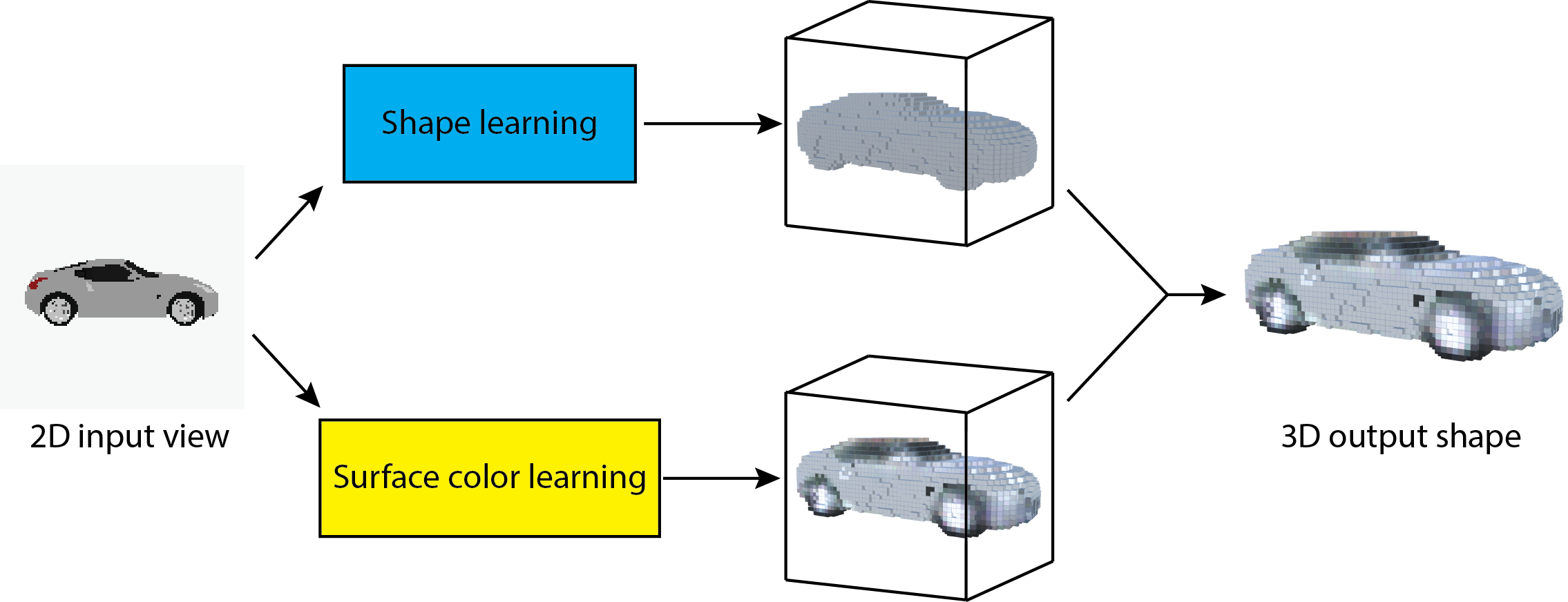}
\centering
\caption{Pipeline of our approach on colorful 3D reconstruction. We decompose the task into two parts: shape reconstruction and surface color estimation. With this framework, our approach is able to achieve visually appealing results.}
\label{fig:teaer}
\end{figure}

Single-image 3D understanding is a long-pursuing goal of computer vision, as a detailed 3D structure recovery can provide valuable information for subsequent cognition tasks like detection and recognition.  
Existing works in this field mainly focus on estimating 3D shape by performing geometric reasoning~\cite{Dyna:SIGGRAPH:2015}\cite{loper2015smpl}\cite{shape_under_cloth:CVPR17}\cite{Bogo:ICCV:2015} or leveraging learning-based priors~\cite{choy20163d}\cite{wu20153d}.
With the emergence of high-capacity deep neural networks~\cite{vgg}\cite{he15deepresidual} and large-scale 3D databases~\cite{chang2015shapenet}, this field has witnessed great progress \textit{w.r.t} both the number and the complexity of the objects that we can model.

However, the visual world around us is not only intrinsically 3D but also colorful.
These two aspects correlate and complement each other, resulting in a comprehensive 3D understanding.
In this work, we push this field a step further by studying a new problem called ``colorful 3D reconstruction'', that is, recovering both the 3D shape and the 3D surface color from a single image, as shown in Figure~\ref{fig:teaer}.
This problem is both intriguing in theory and useful in practice.
For instance, colorful 3D reconstruction provides an automatic way for rapid 3D object prototyping, and greatly extends the functionality of immersive technologies, such as Virtual Reality (VR) and Augmented Reality (AR). 

At the core of the proposed framework are two independent encoding-decoding networks for shape and color learning, respectively.
The shape learning process estimates a 3D geometric shape, and the color learning process predicts surface colors for the estimated shape.
On one hand, we follow conventional shape learning framework by using a shape volume to represent occupancy, but introduce a novel loss function, Mean Squared False Cross-Entropy Loss (MSFCEL), to handle the sparsity problem of volumetric representation.
The introduced MSFCEL balances the loss between occupied and unoccupied voxels, thus enables the model to generate higher resolution 3D models.
On the other hand, we propose a simple but effective concept, color volume, to facilitate the color learning process. 
The color volume is a 3-channel volume with the same spatial dimension as the shape volume. 
Each occupied voxel in the shape volume fetches its color at the same position in the color volume to generate colorful reconstruction results.
While this one-to-one mapping relationship provides an unambiguous way for color sampling, it introduces redundancy for color values of empty voxels.
Such redundant information should be excluded during learning processes, and we achieve this by computing training losses only for surface color voxels.
Our experiments show visually appealing results to prove the effectiveness of the proposed framework.

Our contributions are three-fold: First, we introduce a new problem called ``colorful 3D reconstruction'', which aims to recover both the 3D shape and surface color from a single image.
Second, we propose a novel Colorful Voxel Network (CVN), which combines the strength of appearance hallucination and geometric projection, and integrates them into an end-to-end trainable system.
Finally, we comprehensively evaluate CVN and other state-of-the-art methods across diverse object categories and viewpoints.
We also demonstrate that colorful 3D reconstruction enables many meaningful applications like ``Im2Avatar''.
To our best knowledge, our work is among the first to reconstruct colorful 3D models within the context of deep learning, which we hope can inspire more exciting works.

\section{Related Work}

In this section, we briefly review related work of colorful 3D reconstruction.
3D shape reconstruction has attracted substantial research interests recently due to many vision and graphics applications it promised. 
However, colorful 3D reconstruction is still a challenging and unsolved problem since the 2D-to-3D recovery involves both \textit{satisfying geometric constraints} and \textit{inferring visually appealing textures}. \\

\noindent
\textbf{Geometry-based Reconstruction.}
Geometry-based methods seek to address 3D reconstruction by satisfying geometric constraints.
One line of research focuses on designing scene-dependent 3D models.
For example, in the context of 3D human reconstruction, people have proposed human-specific models like Dyna~\cite{Dyna:SIGGRAPH:2015} and SMPL~\cite{loper2015smpl}.
%
%
Another line of research explores different ways to increase available sensory inputs, such as utilizing multi-view cameras~\cite{shape_under_cloth:CVPR17} or capturing video sequences~\cite{Bogo:ICCV:2015}.
However, for single-view 3D reconstruction, few geometric constraints can be formulated, thus creating great obstacles for existing geometry-based methods.
In this work, we formulate the geometric constraints as differentiable modules in neural networks to enhance their expressive power. \\

\noindent
\textbf{Learning-based Reconstruction.}
Learning-based methods take on the paradigm of data-driven volumetric synthesis.
In particular, the community leverages the recent advances in deep learning to enable powerful modeling of 2D-to-3D mapping.
Large-scale datasets like ShapeNet~\cite{chang2015shapenet} are collected and various neural-network-based approaches are proposed.
For 3D shape estimation, 3DShapeNet~\cite{wu20153d} extended deep belief network (DBN)~\cite{hinton06dbn} to model volumetric shapes;
\cite{choy20163d} unified singe-view and multi-view reconstruction by proposing a 3D recurrent neural network (3D-R2N2)~\cite{choy20163d} with long-short term memory (LSTM)~\cite{lstm}. \cite{zhugwl17} took 2.5D partial volumes as input and the whole architecture is a 3D U-Net\cite{3dunet}. \cite{marrnet} employed 2.5D estimation as an intermediate component and introduced a reprojection consistency loss. \cite{mvcTulsiani18} also discussed the enforced differentiable 2D-3D constraints between shape and silhouettes
For 3D appearance rendering, \cite{tatarchenko2016multi} presented a convolutional encoder-decoder to regress unseen 3D views from a single image.
\cite{zhou2016view} further improved upon it by incorporating a differentiable appearance sampling mechanism. 
%
%
%
Nonetheless, the problem of estimating colorful 3D models from single-view images has not been well studied. 
To this end, we propose a generic framework to concurrently estimate the 3D shape and its surface colors from a single image in an end-to-end manner.
Superior performances are demonstrated for this framework.

\section{Approach}

\begin{figure*}[t]
\includegraphics[width=0.95\textwidth]{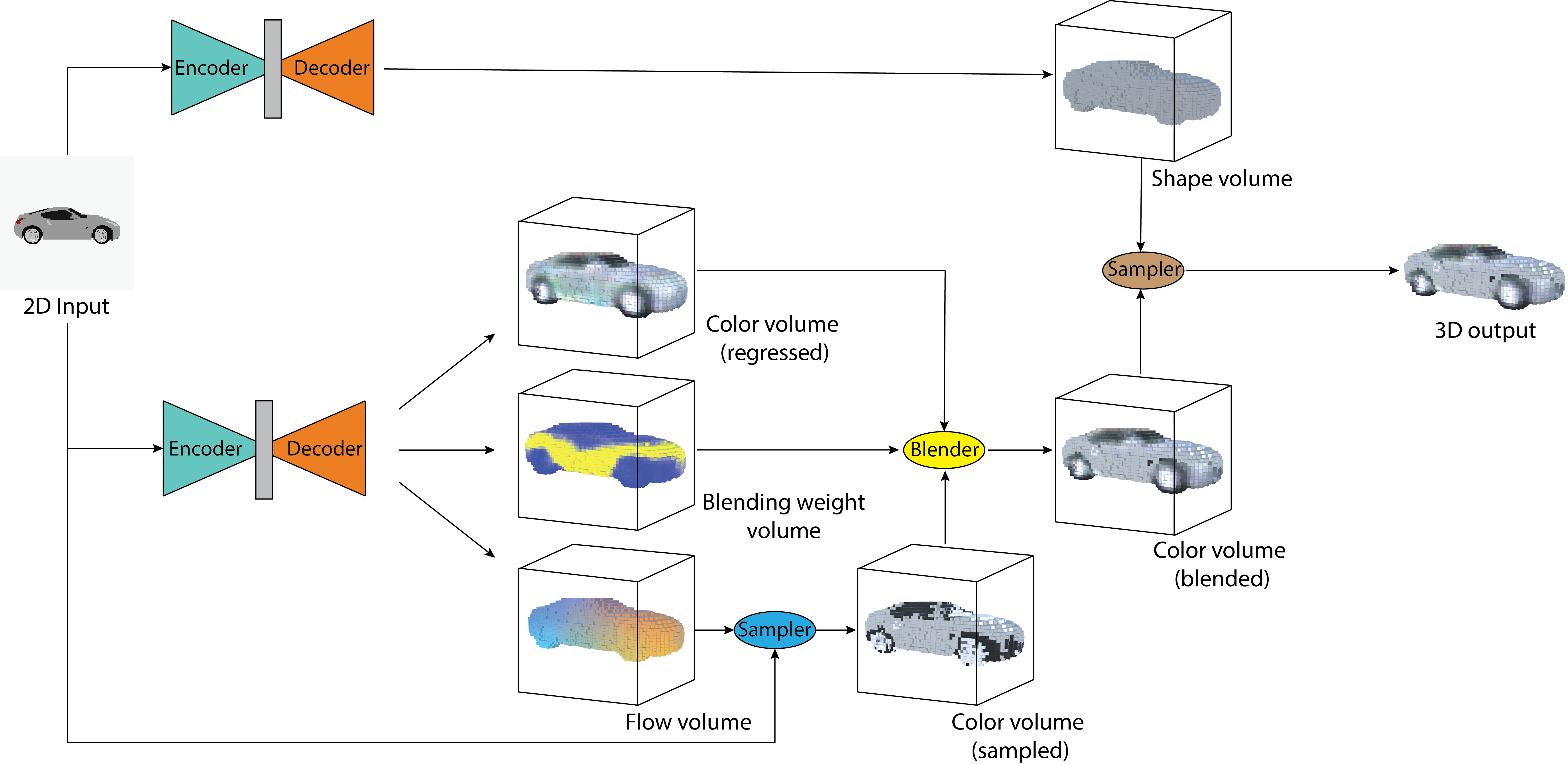}
\centering
\caption{The model architecture used for colorful 3D reconstruction. \textbf{Top branch:} An encoding-decoding network for shape learning. \textbf{Bottom branch:} The pipeline for color learning. }
\label{fig:model}
\end{figure*}

The purpose of this work is to provide an effective framework to estimate a colorful 3D model from a single image. 
We tackle this problem by decomposing shape and color learning into two different procedures, as illustrated in Figure \ref{fig:model}, with each achieved using an encoding-decoding network.
Specifically, the shape learning process outputs an 1-channel volume, with the state of its voxels indicating occupancy. 
The color information is obtained by blending colors generated from two approaches, color regression and color sampling.
Related details are discussed in the following subsections.

\subsection{3D Data Representation}
\label{3d_data_representation}
We first discuss data representation utilized in this work. 
To ease the learning problem, we use volumetric representation, since it is in regular format and compatible with standard CNN models. 
3D volume has been widely used for \textit{shape} reconstruction, but does not receive enough attention for \textit{colorful} 3D reconstruction.
We consider two ways to represent colorful 3D volumes, and we briefly analyze them below.

\noindent
\textbf{Joint Data Representation.}
One way is to jointly represent color and shape as a 3-channel volume. 
This can be achieved by extending the intensity range of color values towards negative to represent empty voxels, while the positive values still represent color in the normal way for occupied voxels. 
For example, we can use the color vector $[1, 0, 0]$ (normalized color intensities) to represent an occupied red voxel, and $[-1, -1, -1]$ for an empty voxel. 
Jointly representing color and shape in this way is spatially efficient, but correlates shape and color learning procedures.
Correlating shape and color affects learning processes in a negative way when the ratio between occupied and unoccupied voxels becomes imbalanced, since the overall color distribution will be pushed towards the color representation of empty voxels. 

\noindent
\textbf{Separate Data Representation.}
The other approach is to represent shape and color separately, for example, using an 1-channel volume for shape and a 3-channel volume for color. 
These two volumes share the same spatial dimensions and are aligned, so that the color and occupancy state of a voxel can be found at the same position within each volume.
This approach is less spatially efficient, since it introduces redundant information (i.e. the color of empty voxels), but it allows us to treat color and shape independently and avoid the learning difficulty due to shape-color correlation. 
We adopt this approach in our work. 
This `shape volume + color volume' scheme is conceptually similar to `mesh + texture image' in GPU rendering framework, in a way that meshes (or shape volumes) provide geometric information and texture images (or color volumes) provide color information for corresponding vertices (or voxels). The main difference is that vertices fetch color through UV mapping, yet surface voxels and their color counterparts are aligned, thus require no explicit mapping. \\

\subsection{3D Shape Learning} 

2D-to-3D shape reconstruction has been widely studied, and most existing models are built on the convolutional encoding-decoding network architecture: given a 2D image $I$, the encoder network $h(\cdot)$ learns a viewpoint-invariant latent representation, which is then fed into a decoder network $g(\cdot)$ to generate a shape volume $\hat{V}$, such that $\hat{V} = g(h(I))$. With the ground truth shape volume $V$ available, the shape reconstruction objective function can be defined using either L2 loss: $\sum_{i} (V_i - \hat{V}_i)^2$ or cross-entropy: $- \sum_{i} [V_i \log{(\hat{V}_i)} + (1 - V_i)\log{(1 - \hat{V}_i)}]$, where $i$ is the voxel index, and $V_i \in \{0, 1\}$ indicating the state of occupancy.

\noindent
\textbf{Handling Sparse Volume.}
We are still on this track for shape reconstruction, but introduce a novel loss function, \textit{Mean Squared False Cross-Entropy Loss} (MSFCEL), to better handle the sparsity problem of volumetric representation.  
MSFCEL is a variant of the \textit{Mean Squared False Error} (MSFE) as proposed in \cite{wang2016training}, and expressed as
\begin{equation}
\label{eq:shape}
MSFCEL = FPCE^2 + FNCE^2
\end{equation}
$FPCE$ is false positive cross-entropy defined on unoccupied voxels of a ground truth shape volume, and $FNCE$ is false negative cross-entropy defined on occupied voxels:
\begin{equation*}
FPCE = -\frac{1}{N}\sum_{n=1}^N [V_n \log{\hat{V}_n} + (1 -V_n)\log{(1 - \hat{V}_n)}]
\end{equation*}
\begin{equation*}
FNCE = -\frac{1}{P}\sum_{p=1}^P [V_p \log{\hat{V}_p} + (1 -V_p)\log{(1 - \hat{V}_p)}]
\end{equation*}
, where $N$ is the total number of unoccupied voxels of $V$, and $P$ is the number of occupied voxels; $V_n$ is the $n^{th}$ unoccupied voxel, and $V_p$ is the $p^{th}$ occupied voxel; $\hat{V}_n$ and $\hat{V}_p$ are the prediction of $V_n$ and $V_p$, respectively.

Equation (\ref{eq:shape}) can be rewritten as $\frac{1}{2}[(FPCE + FNCE)^2 + (FPCE - FNCE)^2]$, thus, minimizing Equation (\ref{eq:shape}) essentially minimizes $(FPCE + FNCE)^2$ and $(FPCE - FNCE)^2$ at the same time. In this way, $MSFCEL$ finds a minimal sum of $FPCE$ and $FNCE$ and their minimal difference concurrently, thus the losses of occupied and unoccupied voxels are minimized together and their prediction accuracies are balanced.

\subsection{Surface Color Learning} 

Inferring complete surface colors of a 3D model given a single view is at the core of visual understanding.
On one hand, this is an ill-posed problem, since occluded surface colors are unaccessible. 
On the other hand, the semantic information available in the input view and some general properties of an object category usually provide clues for human to reasonably infer occluded surface colors. 
For example, if one sees two black tires of a car from its left side, despite this car may have two colorful right tires, he would probably guess the color of its right tires is still black.   
Thus the goal of the proposed color learning framework is to understand such intuition and capture semantic information from a given view for complete surface color estimation.
Two color generation schemes are proposed to jointly achieve this goal: one is sampling colors from the given 2D view, and the other is regressing colors directly. Their results are blended to generate the final surface color.
In our experiments, we found that only considering the voxels corresponding to 3D model surfaces for color loss calculation enhances the learning process and achieves visually pleasant results.

\begin{figure}[t]
\includegraphics[width=0.4\textwidth]{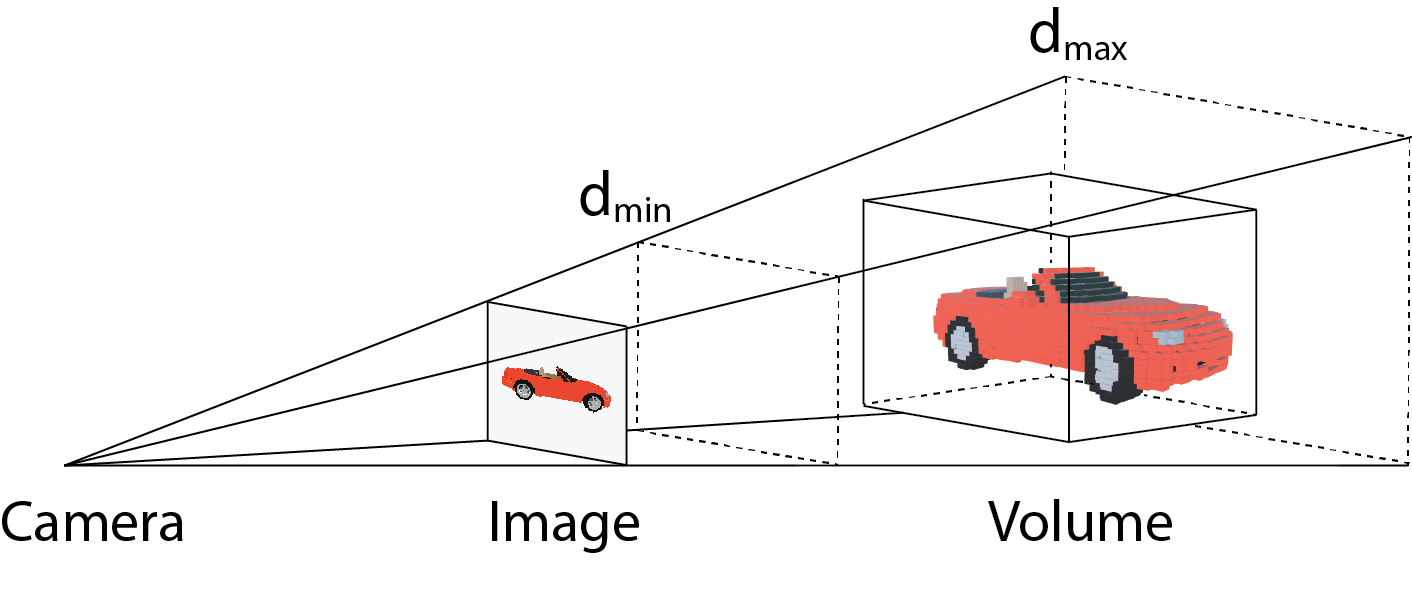}
\centering
\caption{The perspective projection of a camera model.}
\label{fig:projection}
\end{figure}

\noindent
\textbf{3D-to-2D Appearance Flow.}
We name the proposed color sampling scheme as 3D-to-2D appearance flow.
The 2D view of a 3D model usually provides rich semantic information and color sources to inpaint its 3D surface, thus it is a reasonable choice to sample colors from the 2D view.
One way to capture the relationship between the 3D model surface and its 2D view is through perspective projection.
In particular, the 2D projection of a 3D model under a certain camera viewpoint should match the foreground area of its image observation (shown in Figure \ref{fig:projection}).
In other words, given the intrinsic and extrinsic parameters of a camera, one can construct a perspective projection matrix, through which visible surface points are projected onto the 2D view to sample target colors. 
The projected 2D coordinates constitute a 3D-to-2D flow field.
The generated flow field should be smooth to maintain surface texture continuousness.

\begin{figure}[t]
\includegraphics[width=0.4\textwidth]{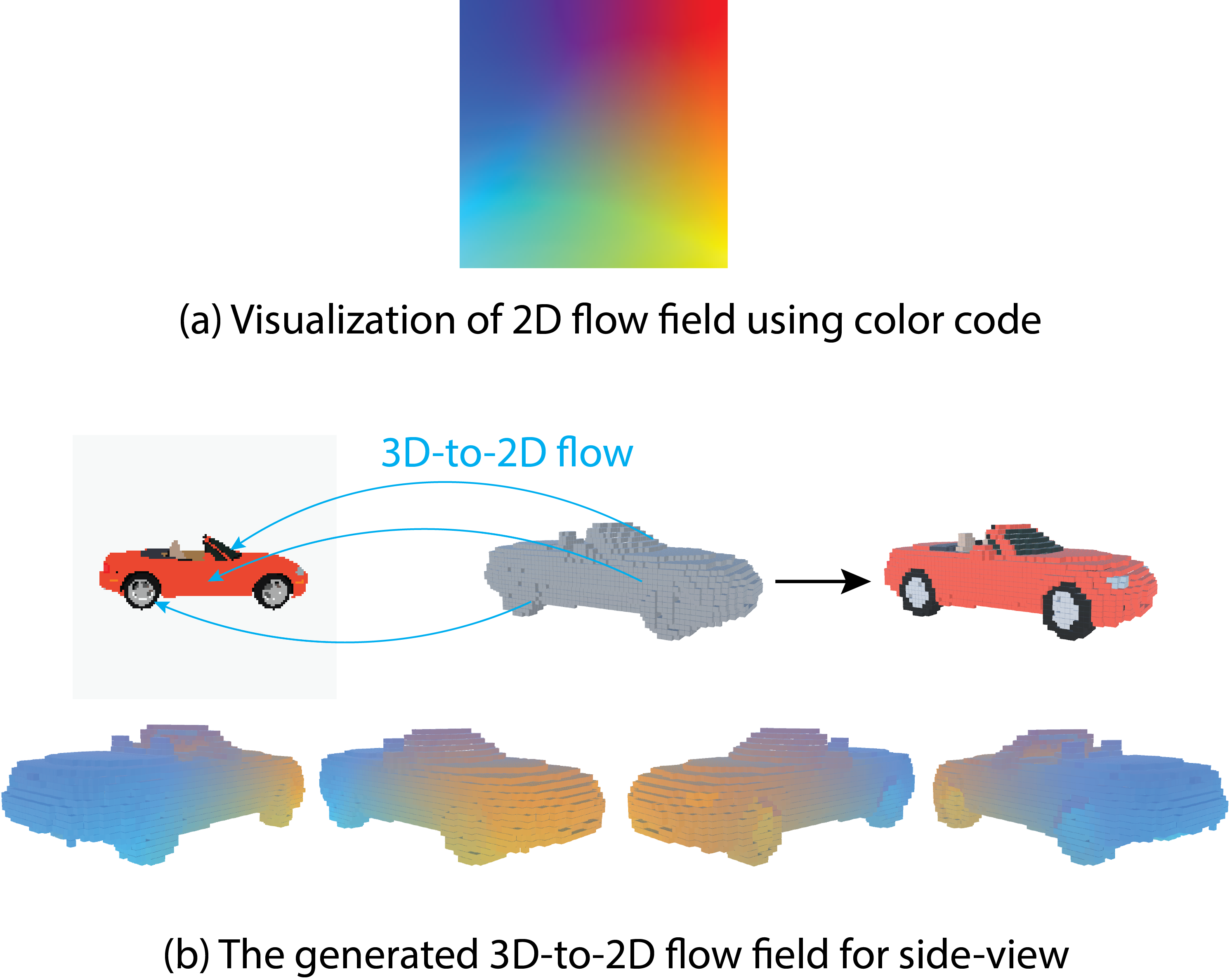}
\centering
\caption{Illustrate the 3D-to-2D appearance flow field. The generated flow fields are visualized from different views in the bottom row of (b), using the color code in (a).}
\label{fig:flow_field}
\end{figure}

Theoretically, a single 2D view does not contain projected coordinates for occluded 3D surface points, but some life experience can help us find reasonable positions to sample colors for them.
For instance, given a side-view of a 3D model, as shown in Figure \ref{fig:flow_field} (b), it is intuitive to sample colors for occluded surface points from the same 2D coordinates as their visible reflectively-symmetric counterparts, resulting in a left-right symmetric flow field.

In our work, we formulate 3D-to-2D appearance flow estimation as a supervised optimization problem.
A 2-channel volume, with each voxel containing a 2D coordinate, is generated to guide color sampling process.
In our work we obtain target flow fields mainly through camera perspective projection, given its 2D view.
The perspective projection is implemented as a $3 \times 4$ transformation matrix $\Theta = K [R \quad T]$, where $K$ is an intrinsic matrix encompassing focal lengths and principal points, and $[R \quad T]$ are extrinsic parameters denoting the coordinate system transformations from 3D world coordinates to 3D camera coordinates.
With $[u\ v\ 1]^T$ representing a 2D point position in pixel coordinates and $[x\ y\ z\ 1]^T$ representing a 3D point position in world coordinates, we can obtain target 3D-to-2D flow vectors, $[u, v]$, for \textit{visible} 3D surface points according to:
\begin{equation}
 \begin{bmatrix}
 u \\
 v\\
 1
 \end{bmatrix}
 \sim
 K [R \quad T]
  \begin{bmatrix}
 x \\
 y \\
 z \\
 1
 \end{bmatrix}
\end{equation}
However, a given 2D view does not contain projected ground truth flow vectors for \textit{occluded} surface points, so the above relationship cannot be directly used for them.
In our experiments, we detect the target flow vector for an occluded 3D surface point by first finding all the foreground 2D points in the view whose colors are similar to that of the considered 3D point, and then selecting the one closest to its projected 2D position via perspective projection.
We assume the perspective transformation matrix is only available during training.
The 3D-to-2D flow vectors are estimated for the whole volume in the forward passing, but only the voxels corresponding to 3D surfaces contribute to loss calculation and error back propagation, because only those voxels are used for surface color rendering.
The objective loss function for 3D-to-2D appearance flow field estimation is defined as:
\begin{equation}
L_{flow} = \frac{1}{S} \sum_{i=1}^S \|V^{flow}_i - \hat{V}^{flow}_i\|_2
\end{equation}
, where $S$ is the number of surface voxels, $V^{flow}_i$ is the target flow vector for the $i^{th}$ surface voxel, and $\hat{V}^{flow}_i$ is its estimation.
Given a 2D view $I$, the sampled color $\hat{V}^{clr\_sample}_i$ using the estimated 2D flow vector for the $i^{th}$ surface voxel is expressed as:
\begin{equation}
\hat{V}^{clr\_sample}_i = \mathbb{S}(I, \hat{V}^{flow}_i)
\end{equation}
, where $\mathbb{S}(,)$ is a 2D sampler (i.e. bilinear sampler).

\noindent
\textbf{Surface Color Regression.}
Regression is a more straightforward way to infer 3D model surface colors, and it can generate colors for positions that are difficult for appearance flow vectors to reach.
The color regression process can be formulated as, given a 2D input view $I$, a 3-channel volume, $V^{color}$, is generated, with each voxel containing a color vector.
In our work, the color intensity is normalized ranging between 0 and 1, thus \textit{sigmoid} activation function is used at the last layer of the network.
We define color regression loss on the surface voxels as: 
\begin{equation}
L_{clr\_regress} = \frac{1}{S}\sum_{i=1}^S \|V^{color}_i - \hat{V}^{clr\_regress}_i \|_2 
\end{equation}
, where $V^{color}_i$ is the color vector of the $i^{th}$ surface voxel, and $\hat{V}^{clr\_reg}_i$ is its estimation using regression.

\noindent
\textbf{Surface Color Blending.}
Based on our observation, the 3D-to-2D appearance flow field can provide visually satisfying color for most smooth surfaces, but often fails for some occluded parts and highly rough regions. 
On the other hand, the regressed surface color tends to be blurry, but can be good alternatives to compensate failure regions of the color sampling scheme.
Thus, it is desired to blend colors from these two process in a way that the overall surface looks visually pleasant to the greatest extent.
To this end, the network generates another 1-channel volume, with each voxel containing a value, $w \in [0, 1]$, to indicate the confidence for the sampled color, and the confidence for the regressed color is $1 - w$.
In our work, we softly blend the two color sources for the $i^{th}$ surface voxel according to:
\begin{equation}
\label{eq:blend}
\hat{V}^{clr\_blend}_i =  w \times \hat{V}^{clr\_sample}_i + (1-w) \times \hat{V}^{clr\_regress}_i
\end{equation}
, and the loss function used to learn the blending weights is defined as:
\begin{equation}
L_{blend} = \frac{1}{S}\sum_{i=1}^S \|V^{color}_i - \hat{V}^{clr\_blend}_i \|_2 
\end{equation}

In summary, the overall objective for surface color learning is:
\begin{equation}
\label{eq:total_loss}
L = L_{flow} + L_{clr\_regress} + L_{blend} 
\end{equation}

\begin{figure}[t]
\includegraphics[width=0.45\textwidth]{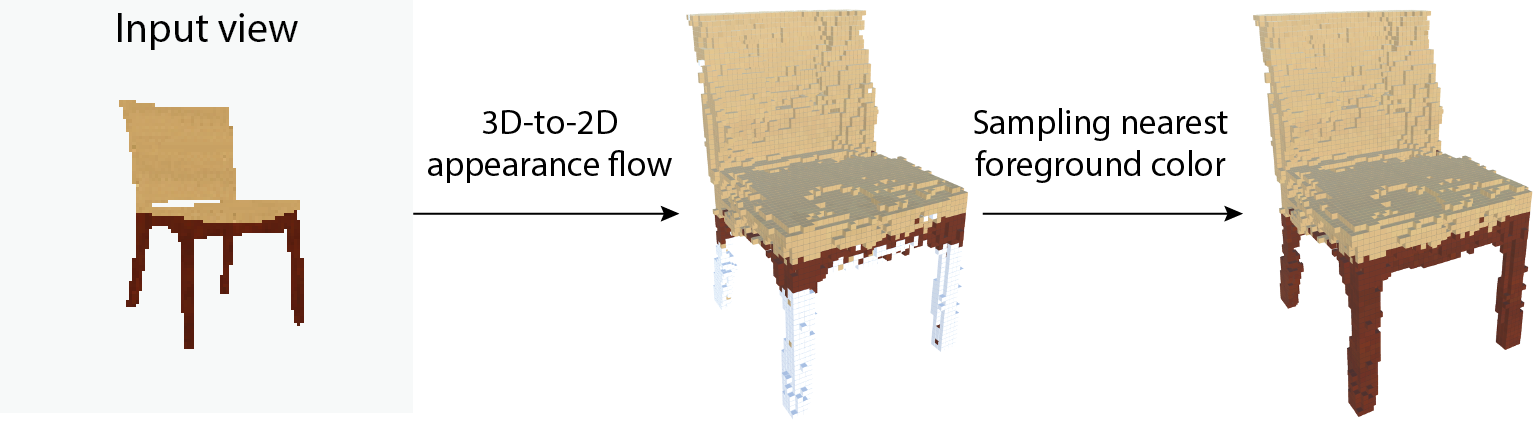}
\centering
\caption{Illustration of the color sampling scheme after estimating appearance flow field. In the middle, color is sampled using estimated flow vectors directly. Note that background color is sampled for chair legs. On the right, color is sampled by detecting the closest foreground pixels to the estimated 2D coordinates.}
\label{fig:nn_sampling}
\end{figure}

\subsection{Implementation Details}

\noindent
\textbf{Network.}
The encoding-decoding networks used for shape and color learning share a similar architecture.
Both encoders contain 6 2D-convolution layers, and can be expressed as $\{(7\times7,64,2), (5\times5,64,2), (5\times5,128,2), (3\times3,128,2), (3\times3,256,2), (4\times4,512,1)\}$ in the format of (filter spatial dimensions, filter channels, stride size).
They take as input a 2D view of size $128\times128$, and generate a 1D latent vector of size $512$.
The latent vector is processed by fully-connected layers and reshaped into size $4\times4\times4\times128$.
Both decoders share the same structure in the first 3 3D-deconvolution layers, represented as $\{(3\times3\times3, 64, 2), (3\times3\times3, 32, 2), (3\times3\times3, 32, 2)\}$, and generate a feature volume of spatial size $32\times32\times32$.
From there, the network for shape learning outputs an 1-channel occupancy volume, while the network for surface color learning outputs 3 volumes: a 3-channel regressed color volume, a 2-channel flow volume and an 1-channel blending weight volume.
All those output volumes are of spatial size $64\times64\times64$.

\noindent
\textbf{Training.}
Since shape and color learning processes are decomposed, they are trained independently. 
The proposed MSFCEL is used as the loss function for shape learning, and Equation (\ref{eq:total_loss}) is used for surface color learning.
At each training step, one view of a 3D model is randomly taken as input to generate a 3D volume.
We retrain the model from the scratch for each category, thus different sets of model parameters are obtained for different categories, respectively. 
Other training details include: optimizer: Adam; learning rate: 0.0003; batch size: 60; training epochs: 500.

\noindent
\textbf{Color Sampling.}
We found that the estimated flow vectors may fail to sample colors for the parts whose 2D projections are thin in the view. 
This is illustrated in Figure \ref{fig:nn_sampling}: background colors are sampled for chair legs when the estimated flow vectors are used directly.
This happens when the shift between the estimated coordinates and the target coordinates is greater than the size of the projected object part.
To achieve better visual effect, the closest foreground pixel to the estimated 2D position is used to sample its color, and one such result is shown on the right side of Figure \ref{fig:nn_sampling}.

\noindent
\textbf{Blending Weight Recalculating.}
The blending weights are used as in Equation (\ref{eq:blend}) during training process.
During testing, we found that assigning more weight than the estimated one to the sampled color is helpful to enhance the visual quality.
In our experiments, we recalculate a new blending weight $w_{new}$ from the estimated one $w_{est}$ according to 
\begin{align*}
 w_{new} &=
  \begin{cases}
   \frac{w_{est}}{\alpha}         & \text{if } w_{est} \leq \alpha \\
   1        & \text{otherwise}
  \end{cases}
\end{align*}
, where $0 < \alpha \leq 1$.  More weights are assigned to sampled colors for smaller $\alpha$; when $\alpha = 1$, the estimated blending weights are used.

\section{Experiments}
\begin{figure}[t]
\includegraphics[width=0.42\textwidth]{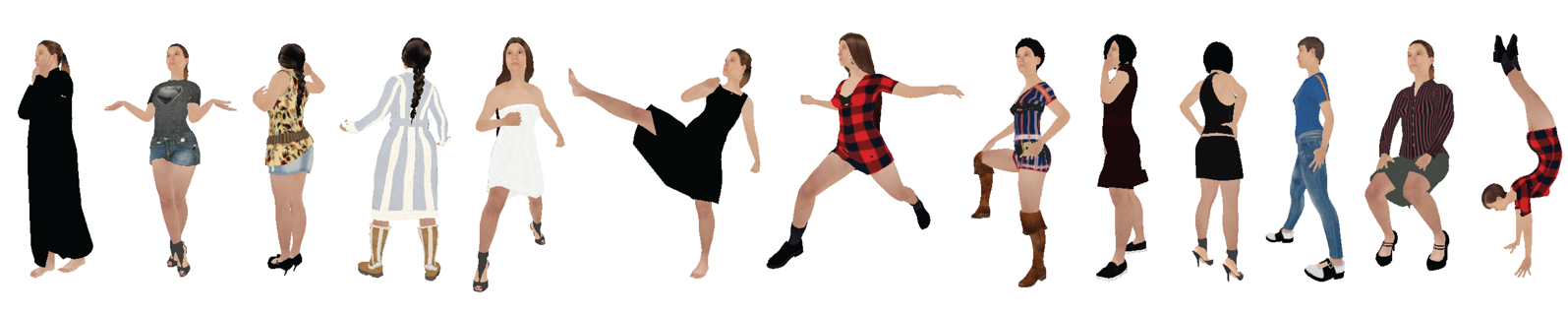}
\centering
\caption{Examples in Colorful Human dataset.}
\label{fig:human_dataset}
\end{figure}

\noindent
\textbf{Datasets.}
To thoroughly investigate the problem of ``colorful 3D reconstruction'', we evaluate on both rigid and articulated objects.
We collect 3D shapes from 4 categories (car, chair, table and guitar) of ShapeNet \cite{chang2015shapenet} for rigid object study.
The 3D shapes in this subset are in mesh format with textures supported.
To align with our pipeline, we first follow \cite{huang1998accurate} to voxelize them into colorful volumes, and then generate 12 2D views for each 3D shape from horizontal viewpoints with an azimuth interval of $30^{\circ}$.
More specifically, 3,389 car models, 2,199 chair models, 2,539 table models and 631 guitar models are collected, and each follows a split ratio of 0.7/0.1/0.2 for training/validation/testing.
To evaluate on articulated objects, we collect 367 3D human models from \textit{MakeHuman}\footnote{\url{www.makehuman.org}}. 
The collected human models (\textit{Colorful Human dataset}) are in mesh format, and examples are shown in Figure \ref{fig:human_dataset}.
Their 2D views and colorful volumes are also generated.
$90\%$ of the processed colorful human shapes are used for training, and $10\%$ for testing in our experiments.

\noindent
\textbf{Evaluation Metrics.}
We first evaluate shape and color quantitatively and separately, and then conduct a user study for qualitative and joint evaluation. Shape is evaluated using Intersection-over-Union (IoU), and color is evaluated using a newly proposed metric, \textit{Surface PSNR}.

IoU is a standard evaluation metric for 3D shape reconstruction \cite{yi2017large}.
For a 3D shape and its prediction, IoU is computed as the ratio between their intersection part and union part.

PSNR is often used as a perceptual quality measurement between the original and a restored image \cite{chen2017trainable}. 
In our experiments, we extend it to evaluate the surface color of reconstructed 3D shapes.
To calculate the proposed surface PSNR, we first align surface voxels between a ground truth shape and its prediction by detecting the closest ground truth surface voxel for each predicted surface voxel (surface voxels belonging to the intersection part of two shapes are detected at the same position); and then calculate PSNR on aligned surface voxels using their color values.

Typically, the higher the two metrics, the better quality of the reconstructed 3D models.
To evaluate the overall colorful 3D reconstruction results (shape and surface color), we also perform a user study to ask test subjects to select the most visually appealing results from different methods.

\noindent
\textbf{Competing Methods.}
We adapt the popular volumetric reconstruction model 3D-R2N2, proposed in \cite{choy20163d}, as the baseline method for performance comparison. 
The 3D-R2N2 model contains a recurrent module, which helps iteratively refine reconstructed 3D volumetric shapes when multiple views are available. 
Since we focus on single-view reconstruction, this recurrent module is dropped. We further adapt the 3D-R2N2 model by adding one more 3D upsampling layer, so that the spatial dimension of the output volume is increased from 32-voxel to 64-voxel to match our framework. 
The output volume is designed to contain 3 feature channels to jointly represent shape and color, as discussed in subsection \ref{3d_data_representation}.
In our experiments, the empty voxel is represented by [-0.5, -0.5, -0.5]. 
During training, L2 loss is used and all voxels contribute to the loss.
All other training settings are kept constant as in our framework.
During testing, a voxel is considered to be nonempty if all its color values are greater than a predetermined threshold, and the color value in each channel is further adjusted according to (using normalized color):
\begin{equation}
\label{eq:clr_adj}
c_{adj} = \frac{c_{pred} - t}{1 - t}
\end{equation}
, where $c_{adj}$ is the adjusted color value, $c_{pred}$ is the predicted color value, and $t$ is the threshold.

\subsection{Evaluation on Rigid Objects}

\begin{figure*}[h!]
\includegraphics[width=0.9\textwidth]{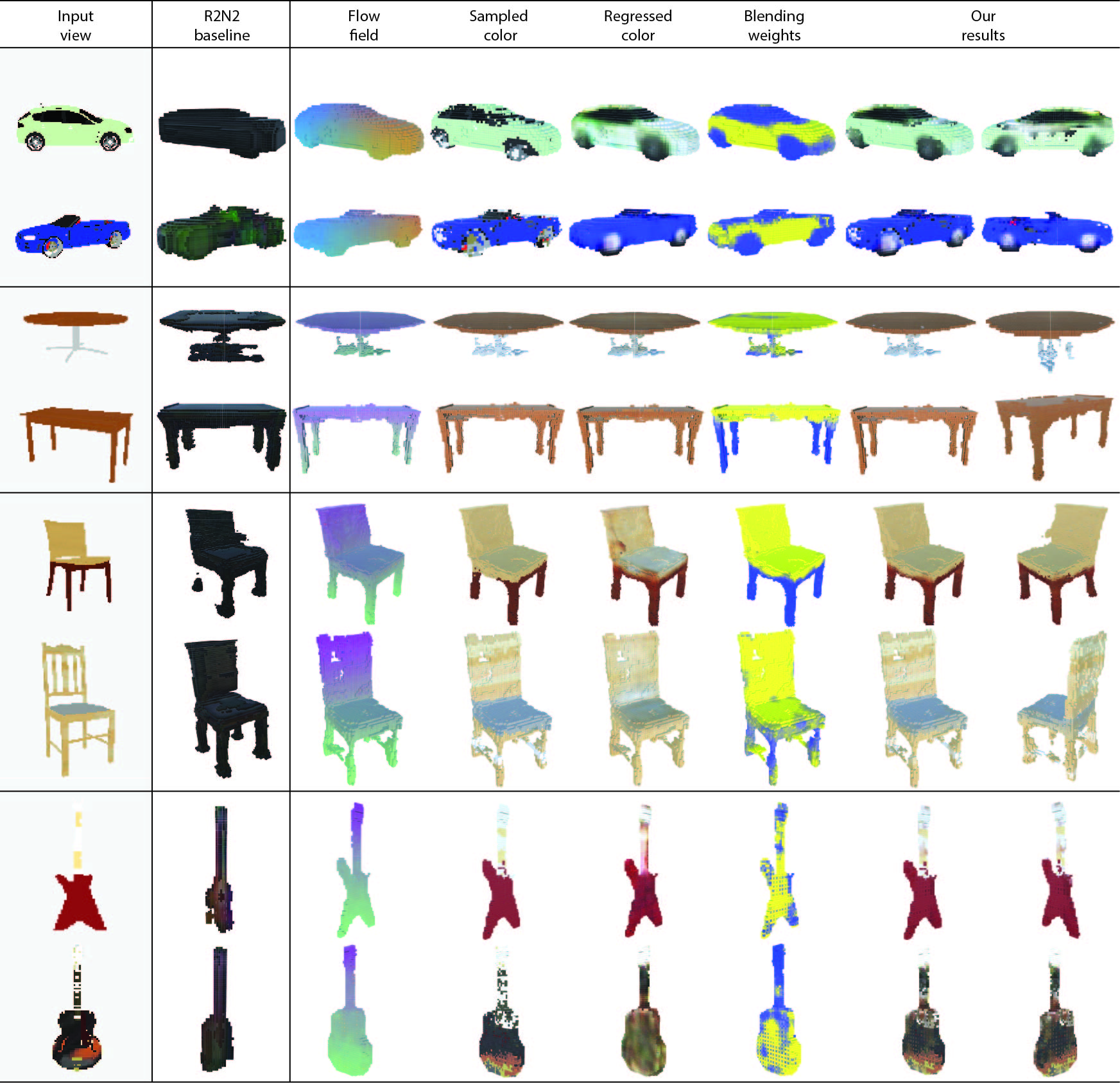}
\centering
\caption{Visualization of intermediate results of the proposed colorful 3D reconstruction framework on testing samples. From left to right (after input view): adapted 3D-R2N2 baseline model, estimated 3D-to-2D appearance flow field (color coded as in Figure \ref{fig:flow_field}(a)), surface color sampled by appearance flow, regressed surface color, blending weights (yellow: high confidence for the sampled color; blue: high confidence for the repressed color), final results under different views.}
\label{fig:res_vis}
\end{figure*}

To fully investigate the proposed learning framework, it is necessary to visualize its intermediate results.
We show testing results in Figure \ref{fig:res_vis}.
For comparison purpose, the results of the adapted R2N2 baseline are also included.
As can be easily noticed, while the R2N2 baseline results present correct geometric structures, their surface colors tend to be blackish.
This can be explained from the fact that 64-voxel volume is sparse and every voxel equally contributes to the loss function, thus the predicted color values are pushed to the color representation of empty voxels, which is [-0.5, -0.5, -0.5].
After color adjustment for occupied voxels, according to Equation \ref{eq:clr_adj}, their color values are close to [0, 0, 0], appearing to be darkish.

For our results, we can observe that the estimated 3D-to-2D appearance flow fields are smooth, and approximately reflect reasonable color sampling positions in 2D views.
A closer look reveals that the sampled colors are satisfying for smooth regions (both visible and occluded), yet fails for highly rough areas (i.e. tires and windows of cars) and thin shape parts (i.e. table and chair legs).
On the other hand, the regressed surface colors are blurry and loses detailed textures, but can be used to compensate for the failed color sampling regions.
It is also shown that the predicted weights for blending those two color sources can reflect such compensation properly and achieve visually appealing results.

\subsection{Ablation Study}
We conduct ablation study on ShapeNet subset to evaluate shape and surface color learning procedures in this subsection.

\noindent
\textbf{The Effectiveness of MSFCEL.}
The introduced MSFCEL is designed to better handle the sparsity problem in the process of 3D shape reconstruction.
To evaluate its effectiveness, we compare it with standard cross-entropy loss function.
More specifically, we retrain the same network for 3D shape reconstruction after only changing the loss function from MSFCEL to cross-entropy while keeping other settings constant.
All the models are trained independently for each shape category, and generate 12 3D shapes for the 12 views of each testing instance. 
We use mean IoU for evaluation, which is calculated by averaging the IoU of all generated 3D shapes in a category.
The results are reported in Table \ref{table:meaniou}, with those of the adapted 3D-R2N2 baseline model included.
Note that, Our model trained using MSFCEL outperforms the other two models for all shape categories.
Both our shape reconstruction model and the adapted R2N2 baseline model contain $\sim{7}$M trained parameters.
The R2N2 baseline model may achieve better shape reconstruction results if it is exclusively constructed for shape learning.

\begin{table}[t]
 \scriptsize
 \caption{Mean IoU on testing samples}
 \centering
 \label{table:meaniou}
 \begin{tabular}{c c c c c | c}

 {} & car & table & chair & guitar & avg.\\
 \hline
 Adapted R2N2 & 0.238 & 0.216 & 0.179 & 0.308 & 0.235\\
 Ours (Cross-entropy) & 0.381 & 0.226 & 0.172 & 0.342 & 0.280 \\
 Ours (MSFCEL)  & \textbf{0.395} & \textbf{0.241} &  \textbf{0.180} & \textbf{0.374} & \textbf{0.298}\\
 \hline
 \end{tabular}
\end{table}

\noindent
\textbf{The Effectiveness of Selective Blending.}
The blending scheme proposed in our framework aims to compensate the two color sources estimated from the same network, and achieve visually appealing results.
To evaluate its effectiveness, we compare it with two baseline models.
One is \textit{color regression model}, which is obtained by dropping the network branches that generate the blending weight volume and the flow volume (and related color sampling operations) in surface color learning process. 
The other one is \textit{flow model}, which is obtained by only keeping the flow volume and its color sampling operation.
The loss function used for training the color regression model and the flow model only contains $L_{clr\_regress}$ and $L_{flow}$, respectively.
We also evaluate the influence of different choices of $\alpha$ (used for recalculating blending weights) on the visual quality of our results.
The $\alpha$ is uniformly sampled from $0.2$ to $1.0$ with an interval of $0.2$ in our experiment.
Still, all the models are trained category-independently and generate 12 3D shapes for each testing instance. 
The introduced metric, surface PSNR, is used to evaluate the perceptual quality of estimated surface colors.
We calculate surface PSNR in both RGB and YCbCr color spaces for all the reconstructed shapes, and their mean values are reported for each category in Table \ref{table:psnr_rgb} and Table \ref{table:psnr_ycc}.
The adapted R2N2 baseline model performs worst since it generates overall darkish color as shown in Figure \ref{fig:res_vis}.
In both tables, the color regression model performs best in car category, and worse than the flow model for table and chair.
Our model with color blending scheme ($\alpha=0.2$) achieves the best results for table, chair, guitar and average, indicating the effectiveness of the proposed selective blending scheme.

 \begin{table}[h!]
  \caption{Mean surface PSNR (RGB) on testing samples}
  \scriptsize
  \label{table:psnr_rgb}
  \begin{tabular*}{0.47\textwidth}{c c c c c | c}
  {} & car & table & chair & guitar & avg. \\
  \hline
  Adapted R2N2 & 6.502 & 6.901 & 7.063 & 5.668 & 6.533\\
  Our baseline (clr. reg.) & \textbf{11.005} & 17.621 & 16.715 & 10.598 & 13.984 \\
  Our baseline (flw.)  & 8.820 & 17.651 & 17.444 & 9.825 & 13.435 \\
  Ours ($\alpha$ = 0.2) & 10.881 & \textbf{19.728} & \textbf{18.252} & \textbf{10.617} & \textbf{14.870} \\
  Ours ($\alpha$ = 0.4) & 10.924 & 18.696 & 17.965 & 10.572 & 14.539\\
  Ours ($\alpha$ = 0.6) & 10.914 & 18.724 & 17.800 & 10.543 & 14.495 \\
  Ours ($\alpha$ = 0.8) & 10.902 & 18.175 & 17.659 & 10.522 & 14.314 \\
  Ours ($\alpha$ = 1.0) & 10.892 & 18.278 & 17.517 & 10.507 & 14.298 \\
  \hline
  \end{tabular*}
 \end{table}

\begin{table}[h!]
 \caption{Mean surface PSNR (YCbCr) on testing samples}
 \scriptsize
 \label{table:psnr_ycc}
 \begin{tabular*}{0.47\textwidth}{c c c c c | c}
 {} & car & table & chair & guitar & avg. \\
 \hline
 Adapted R2N2 & 11.300 & 11.342 & 11.494 & 10.287 & 11.106\\
 Our baseline (clr. reg.) & \textbf{15.524} & 22.262 & 21.189 & 15.251 & 18.557 \\
 Our baseline (flw.)  & 13.289 & 22.305 & 21.951 & 14.523 &  18.017 \\
 Ours ($\alpha$ = 0.2) & 15.399 & \textbf{24.374} & \textbf{22.750} & \textbf{15.270} & \textbf{19.448} \\
 Ours ($\alpha$ = 0.4) & 15.448 & 23.341 & 22.458 & 15.218 & 19.116\\
 Ours ($\alpha$ = 0.6) & 15.441 & 23.357 & 22.290 & 15.185 & 19.068 \\
 Ours ($\alpha$ = 0.8) & 15.430 & 22.813 & 22.146 & 15.162 & 18.888 \\
 Ours ($\alpha$ = 1.0) & 15.420 & 22.904 & 22.002 & 15.145 & 18.868 \\
 \hline
 \end{tabular*}
\end{table}

\begin{figure}[h!]
\includegraphics[width=0.45\textwidth]{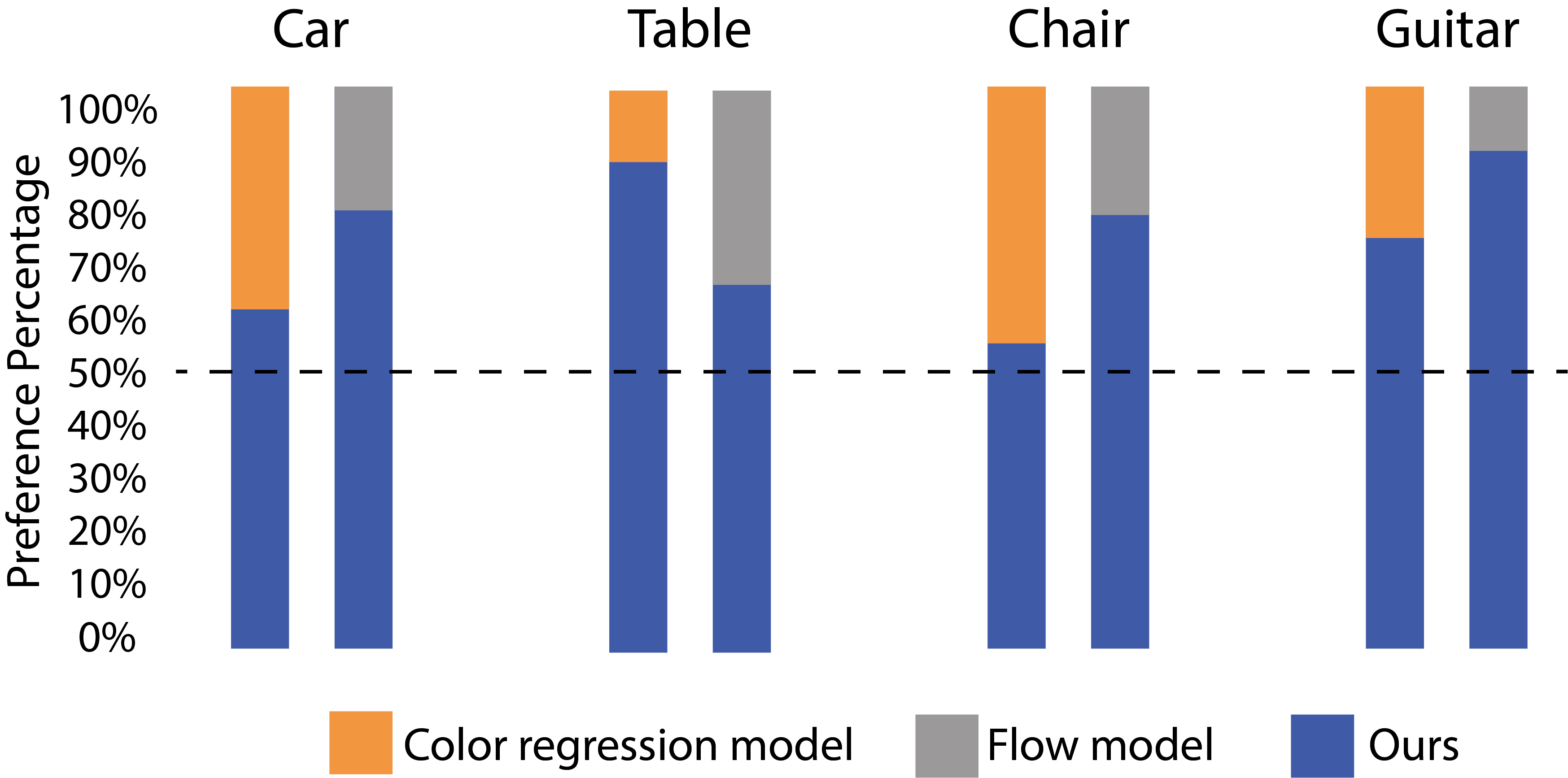}
\centering
\caption{User study results of our results against our baseline models: color regression model and flow model. The preference percentages for our results are plotted (blue), and the results of baseline models are plotted as their compensation parts (orange and gray).}
\label{fig:user_study}
\end{figure}

We conduct a user study to evaluate the overall quality of colorful 3D reconstructed results.
The user study includes 20 randomly selected 3D shapes, 5 for each category.
Each instance contains 3D reconstruction results of color regression model, flow  model and our model ($\alpha=0.2$), together with their input 2D views.
All results are grouped into pairs in the format of (our result, color regression/flow result).
10 test subjects with no computer vision experience are asked to select their preferences on each pair.
We report their average preference percentages for our results in Figure \ref{fig:user_study}.
Our results receive higher preferences over all shape categories, which also indicates the effectiveness of the proposed blending scheme.

\begin{figure}[h!]
 \includegraphics[width=0.48\textwidth]{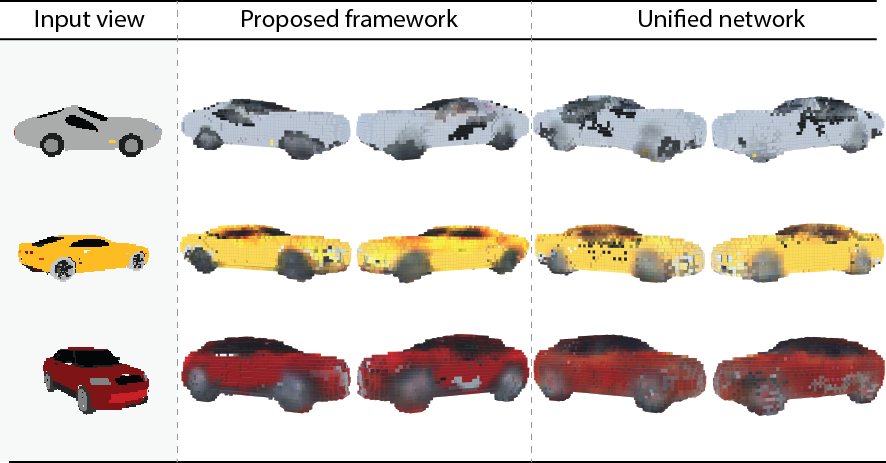}
 \centering
 \caption{Visualize \textit{randomly} selected colorful reconstruction results of the proposed framework and the unified network. The 3D shapes are rendered from two different views.}
 \label{fig:model_str_ana}
 \end{figure}

\noindent
\textbf{Model Structure Analysis.}
The proposed framework decomposes shape and color learning processes by using two independent encoding-decoding networks.  In this part, we experiment with a unified network architecture for jointly learning shape and surface colors, and compare its performance with the proposed one.
The unified network generates all the 4 volumes from only one encoding-decoding network.
In our experiment, we implement it by reusing the network architecture for surface color learning, and adding one more output branch for shape volume.
The loss function is changed to $L_{flow} + L_{clr\_regress} + L_{blend} + MSFCEL$, and other settings are kept constant.
We trained and tested the unified network on car category.
The testing mean IoU for the unified network model is $0.386$, which is less than $0.395$ (for the proposed framework). 
We also compare \textit{randomly} selected colorful reconstruction results of both pipelines in Figure \ref{fig:model_str_ana}.
As can be observed, the surface colors generated by our proposed framework are smoother and more visually satisfying.

\subsection{Evaluation on Articulated Objects}

\begin{figure}[h!]
\includegraphics[width=0.46\textwidth]{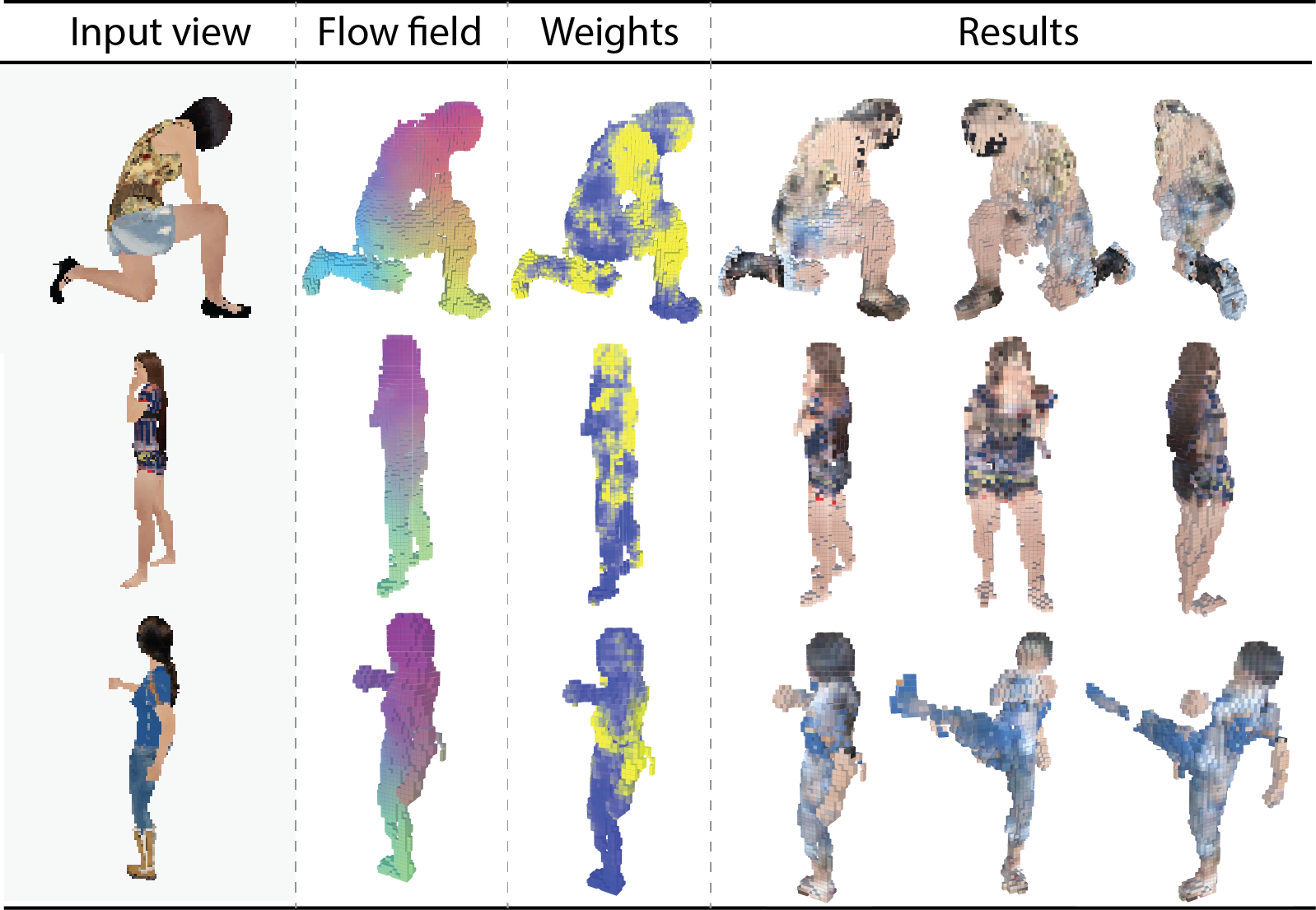}
\centering
\caption{Visualize colorful reconstruction results of Colorful Human Dataset.}
\label{fig:human_res}
\end{figure}

We further test the proposed framework on Colorful Human Dataset. 
The human models present more asymmetric geometric properties than ShapeNet CAD models, and usually contain more colorful textures, which increase the difficulty for colorful 3d reconstruction. 
Yet, our proposed framework can still generate visually pleasant results.
We visualize 3D-to-2D appearance flow fields, blending weights and our final results in Figure \ref{fig:human_res}.
Note that, the estimated flow fields are overall smooth and roughly reflect correct color sampling positions.
Also, the blending weights are able to properly blend color sources, such that occluded human body parts can still be reasonably estimated (i.e. face).

\begin{figure}[h!]
\includegraphics[width=0.48\textwidth]{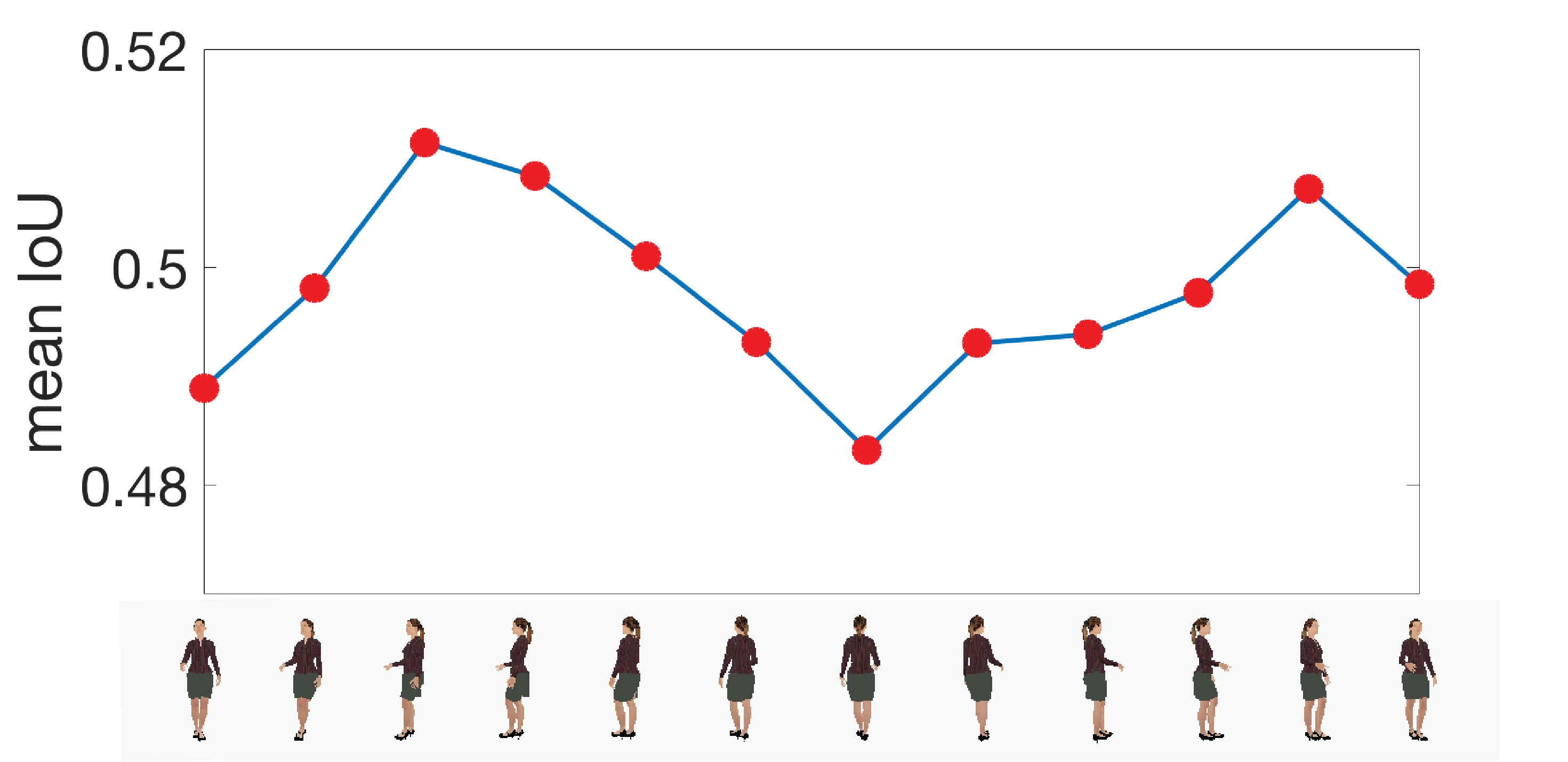}
\centering
\caption{Mean IoU for different views.}
\label{fig:human_iou_view}
\end{figure}

\begin{figure}[h!]
\includegraphics[width=0.32\textwidth]{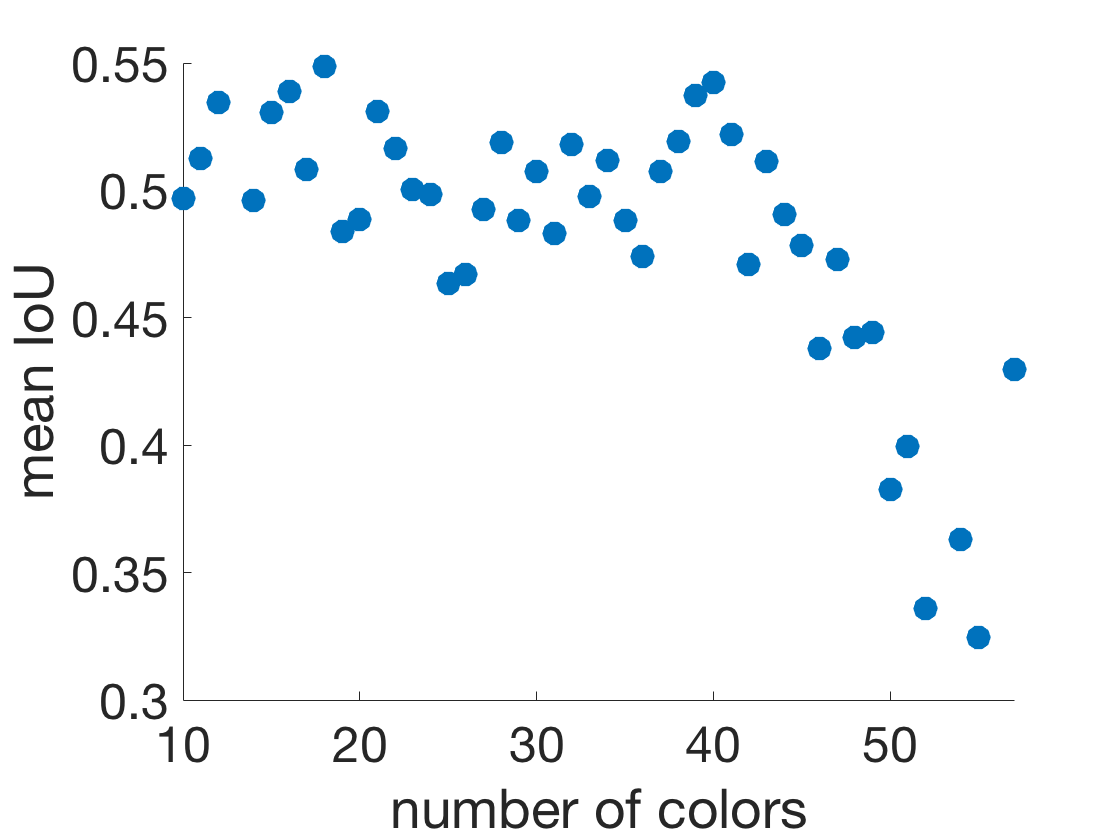}
\centering
\caption{Mean IoU for different color complexities.}
\label{fig:human_iou_clr}
\end{figure}

\subsection{Further Analysis}

\noindent
\textbf{The Influence of Geometric Viewpoints.}
Since human shapes present asymmetric properties, we study the effect of geometric viewpoints on 3D shape reconstruction.
The mean IoU is calculated for different views, and plotted in Figure \ref{fig:human_iou_view}.
As shown, compared to front and back views, side views generate better reconstructed human shapes.
This can be explained from the fact that human bodies are usually left-right symmetric but not front-back symmetric, and side views also contain more complete details.

\noindent
\textbf{The Influence of Color Complexity.}
Human models usually contain more colorful textures, thus we investigate the influence of \textit{color complexity} on 3D shape reconstruction in this part.
Here, we define color complexity as the number of colors in the input view, and calculate it as follows.
We first summarize all the unique colors and their counts from the whole Colorful Human Dataset, and filter out the colors whose counts are less than a predetermined number, $t_1$.
The retained colors are clustered using $K$-means.
Then, for each input view, we assign all its foreground colors to their closest cluster centers, and count the number of cluster centers whose assignments are greater than a threshold $t_2$ as the color complexity.
In our experiment, we set $t_1 = 50$, $K = 128$, and $t_2 = 5$.
We plot the mean IoU for different color complexities in Figure \ref{fig:human_iou_clr}.
The mean IoU maintains high for color complexities less than around $40$, and drops when color complexity increases.

\section{Conclusion}

In this work, we introduce a new ``colorful 3D reconstruction'' task and present a general framework, Colorful Voxel Network (CVN), to tackle this less studied problem. 
CVN learns to decompose comprehensive 3D understanding into two subproblems: 3D shape learning and surface color learning. 
Specifically, for color learning, a unified network is proposed to generate color by combining the strengths of two different approaches (regression-based hallucination and flow-based sampling), and adaptively blends them in a visually pleasant way.
Extensive experimental results demonstrate the effectiveness of the proposed framework and also illustrate that it enables many meaningful applications like ``Im2Avatar''.

{\small
\bibliographystyle{ieee}
\bibliography{egbib}
}

\end{document}